\documentclass[10pt,twocolumn,letterpaper]{article}

\usepackage[pagenumbers]{cvpr}

\usepackage{graphicx}
\usepackage{amsmath}
\usepackage{booktabs}
\usepackage[stretch=10,shrink=10]{microtype}
\usepackage{units}
\usepackage{subcaption}
\usepackage{caption}

\makeatletter

\def\cvprsection{\@startsection{section}{1}{\z@}
   {-8pt plus -2pt minus -2pt}{4pt}{\large\bf}}
\def\cvprsubsection{\@startsection{subsection}{2}{\z@}
   {-4pt plus -2pt minus -2pt}{3pt}{\elvbf}}
\makeatother
\renewcommand{\paragraph}[1]{\vspace{.25em}\noindent\textbf{#1}~}
\definecolor{cvprblue}{rgb}{0.21,0.49,0.74}
\usepackage[pagebackref,breaklinks,colorlinks,allcolors=cvprblue]{hyperref}

\def\paperID{*****}
\def\confName{CVPR}
\def\confYear{2026}

\title{DENSER: Depth-Guided Ensemble with Staged EFA-GS Reconstruction for Soccer Novel View Synthesis}

\author{Parthsarthi Rawat\\
GameChanger by Dick's Sporting Goods\\
{\tt\small sarthi.rawat@gc.com}
}

\begin{document}
\maketitle

\begin{abstract}
We propose \textbf{DENSER}, a Depth-guided ENSemble with Staged EFA-GS
Reconstruction for soccer novel view synthesis. DENSER extends
EFA-GS with three key contributions: (1)
camera-height-based loss weighting that prioritises ground-level broadcast
views, (2) monocular depth supervision from Depth-Anything-V2 to regularise
geometry in textureless regions, and (3) a three-model pixel-average ensemble
whose members diverge from a shared base checkpoint by varying training length
and Gaussian scale clamping. On five held-out challenge scenes we achieve a mean
PSNR of \textbf{29.89\,dB}, SSIM of \textbf{0.791}, and LPIPS of \textbf{0.366}.
\end{abstract}

\section{Method}

\subsection{Base Model}
Our base representation is \textbf{EFA-GS}~\cite{wang2025efags}, which builds
on Mip-Splatting~\cite{Yu2023MipSplatting}, an alias-free variant of
3D Gaussian Splatting~\cite{kerbl3Dgaussians} that replaces the 2D
screen-space dilation filter with a 3D smoothing kernel to suppress aliasing.
EFA-GS further introduces low-frequency-first initialisation and progressive
scale annealing to eliminate floating Gaussian artefacts that afflict standard
training. All models use spherical harmonic degree~3, a 3D smoothing kernel size of~0.1,
and are trained on half-resolution images ($2048\times1080$).

\subsection{Camera-Height Weighting}
\label{sec:camweight}
The SoccerNet-NVS dataset~\cite{snNVS} spans cameras at vastly different heights:
ground-level views capture the most perceptually important angles yet are
outnumbered by overhead cameras, causing unconstrained training to underfit them.
We weight each camera's photometric loss by altitude band. Each scene contains
${\approx}420$ training cameras classified by world-space Y, where Y decreases
with physical height following the COLMAP convention (Table~\ref{tab:camweights}).
Weights are normalised by their mean so the effective learning rate is unchanged.
\begin{center}
\small
\setlength{\tabcolsep}{4pt}
\begin{tabular}{lcc}
\toprule
Band & Y range & Weight \\
\midrule
Ground    & $Y > -5$          & 5.0 \\
Sideline  & $-12 < Y \le -5$  & 3.0 \\
Mid       & $-20 < Y \le -12$ & 3.0 \\
Overhead  & $Y \le -20$       & 1.0 \\
\bottomrule
\end{tabular}
\captionof{table}{Camera altitude bands and photometric loss weights.}
\label{tab:camweights}
\end{center}

\subsection{Depth Supervision}
\label{sec:depth}
Soccer scenes contain large textureless regions — grass, sky, and stadium
stands — where photometric loss alone provides weak gradient signal for
Gaussian placement. To regularise geometry in these regions we supervise
training with monocular pseudo-depth maps generated offline using
\textbf{Depth-Anything-V2-Large}~\cite{depth_anything_v1,depth_anything_v2}
(ViT-L encoder). A scale-and-shift invariant $L_1$ loss is applied every
5 iterations after iteration 1\,000:
\begin{equation}
\mathcal{L}_\text{depth} = \lambda_\text{depth} \cdot
  \min_{s,t} \| s \cdot \hat{D} + t - D_\text{gt} \|_1
\end{equation}
where $\hat{D}$ is the rendered depth and $D_\text{gt}$ is the pseudo-depth;
scale-and-shift invariance accounts for the metric ambiguity of monocular
estimates.

\begin{figure*}[t]
\centering
\begin{subfigure}[b]{0.325\linewidth}
  \includegraphics[width=\linewidth]{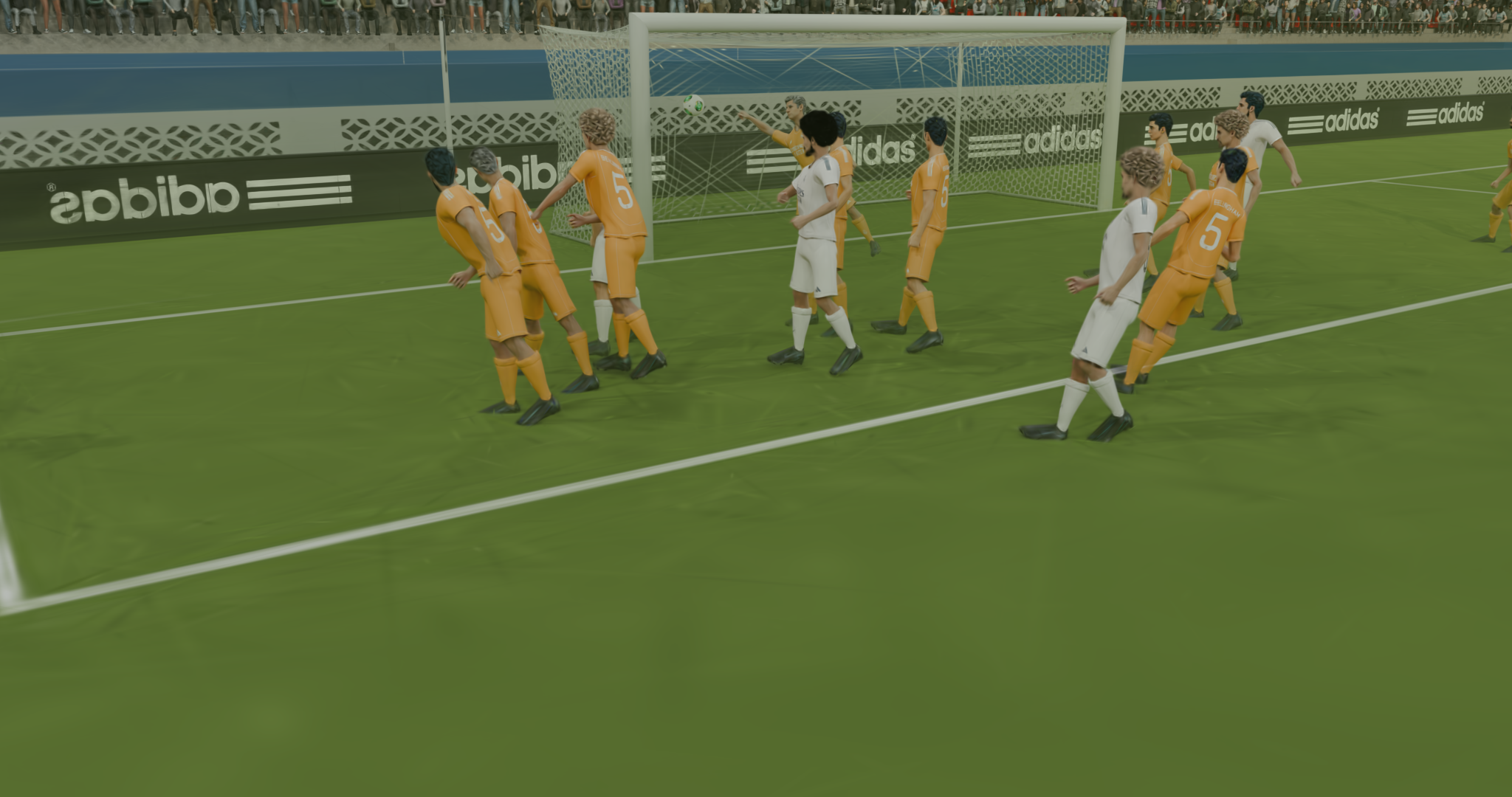}
  \caption{3DGS (baseline)}
\end{subfigure}
\hfill
\begin{subfigure}[b]{0.325\linewidth}
  \includegraphics[width=\linewidth]{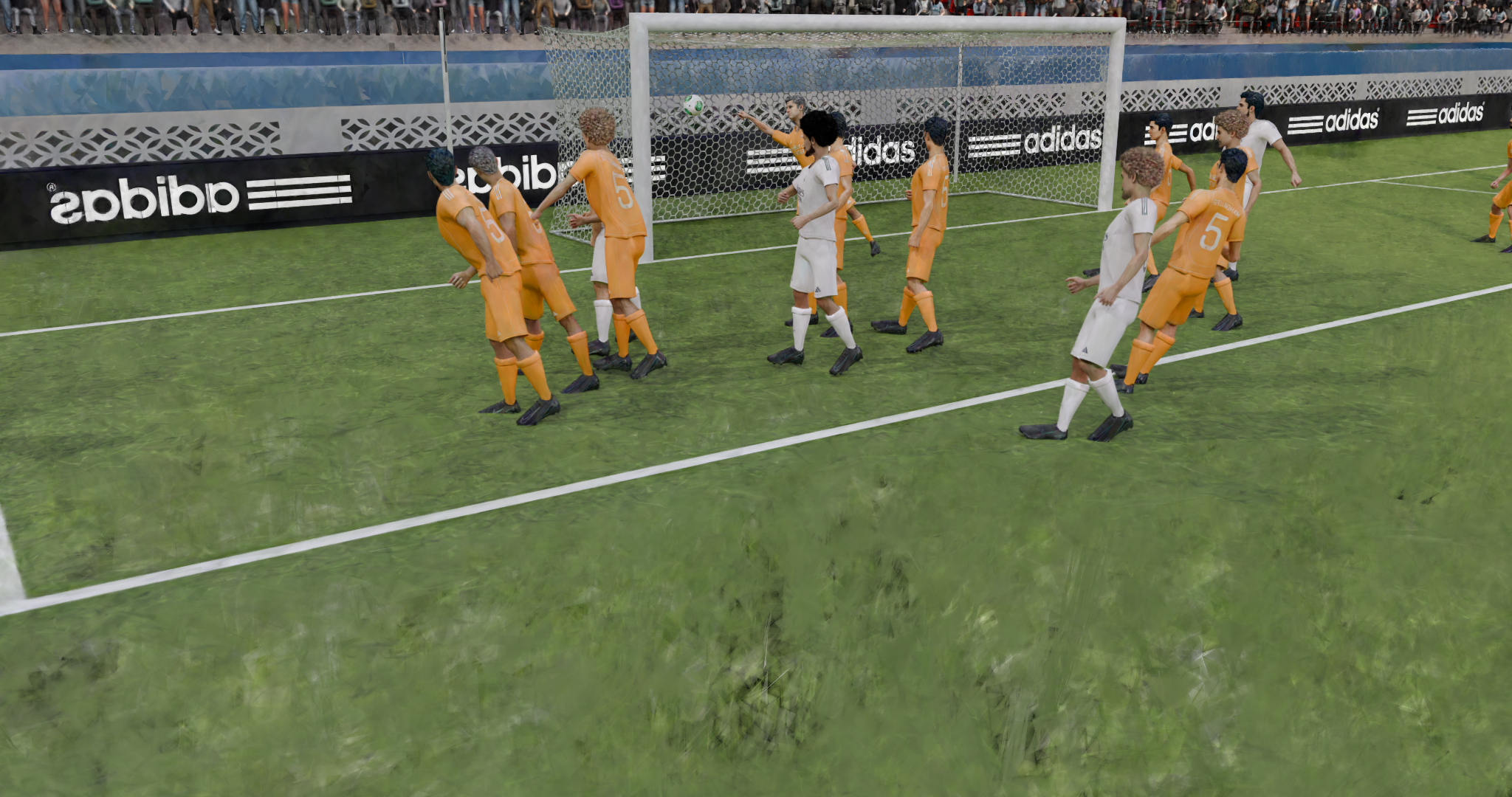}
  \caption{Triangle Splat (baseline)}
\end{subfigure}
\hfill
\begin{subfigure}[b]{0.325\linewidth}
  \includegraphics[width=\linewidth]{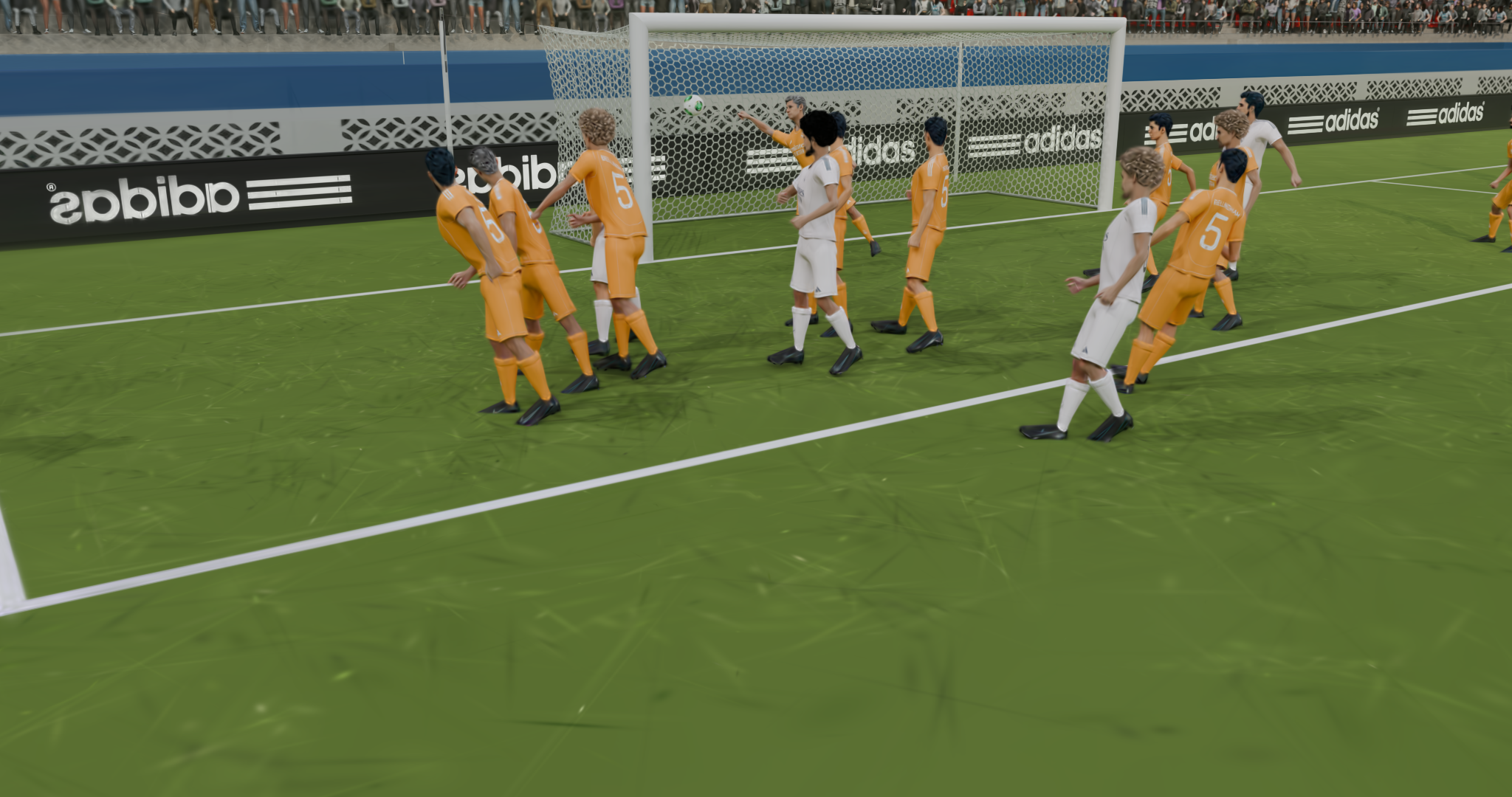}
  \caption{DENSER (Ours)}
\end{subfigure}
\caption{Qualitative comparison on Scene~3, camera~22. 3DGS exhibits a pervasive green colour cast and loses fine structure on the goal net. Triangle Splat produces visible triangular surface artefacts across the pitch and washed-out colours. DENSER is free of all these failure modes.}
\label{fig:comparison}
\end{figure*}

\subsection{Staged Training Pipeline}
Training proceeds in two shared stages that build a strong geometric foundation,
followed by three diverging fine-tuning branches that differ in training duration
and Gaussian scale constraints. Branching from one shared checkpoint ensures the ensemble members
are maximally diverse in their later-stage errors while sharing the same
well-initialised geometry.

\paragraph{Stage~1: Geometry Initialisation (0--50k iterations).}
A fresh EFA-GS model is trained from scratch for 50\,k iterations with
camera-height weighting applied throughout all training. Gaussian scale annealing
(initial max scale $2.0 \to 1.0$) allows primitives to start large enough to
cover broad scene regions and progressively specialise as training converges.
Depth supervision is deliberately withheld at this stage: applying it to a
randomly-initialised scene causes pseudo-depths to anchor Gaussians at
incorrect positions before the photometric signal has established coarse
geometry.

\paragraph{Stage~2: Depth-Guided Consolidation (50--90k iterations).}
Once coarse geometry has converged, depth supervision is introduced
($\lambda_\text{depth}=0.05$). The stronger signal corrects remaining depth
ambiguities — particularly in textureless grass regions where photometric
gradients alone are weak. A short densification pass (iterations
50\,001--55\,000, gradient threshold $\tau=0.0002$) fills under-reconstructed
regions exposed by the depth signal. A shared checkpoint is saved at 90\,k;
all three Stage~3 branches start from this point.

\paragraph{Stage~3a: Short Fine-tuning (90--110k iterations).}
The shared checkpoint is continued for 20\,k more iterations with reduced depth
supervision ($\lambda_\text{depth}=0.01$). The lower weight is necessary
because a second densification pass (90\,001--97\,000, $\tau=0.0002$,
minimum opacity $0.01$) spawns new Gaussians that are still photometrically
immature; a strong depth signal at this point would anchor them to incorrect
positions.

\paragraph{Stage~3b: Extended Training with Scale Clamping (90--130k iterations).}
Identical to Stage~3a but extended to 130\,k iterations with a hard scale
clamp (maximum Gaussian scale $2.0$) applied at every densification step.
Without the clamp, longer training allows a small number of Gaussians to grow
into large elongated ellipsoids that project as visible colour streaks across
advertising boards and pitch boundaries.

\paragraph{Stage~3c: Extended Training without Scale Clamping (90--130k iterations).}
Identical to Stage~3b ($\lambda_\text{depth}=0.01$, 130\,k iterations) but without the scale clamp. Unconstrained Gaussians fit large homogeneous regions more faithfully at
the risk of occasional streak artefacts. Stage~3b and Stage~3c therefore
capture complementary error profiles: averaging them suppresses streaks while
retaining the reconstruction quality of the unconstrained model.

\subsection{Ensemble}
We combine the three Stage~3 models by rendering each independently and
pixel-averaging with equal weights:
\begin{equation}
I_\text{ens} = \tfrac{1}{3}\bigl(I_\text{3a} + I_\text{3b} + I_\text{3c}\bigr).
\end{equation}
Stage~3a is shorter-trained and strongly regularised by depth; Stage~3b
suppresses large-scale artefacts via scale clamping; Stage~3c benefits from
longer unconstrained optimisation. Pixel averaging reduces per-model noise and
partially cancels the systematic errors unique to each branch.

\section{Implementation Details}

\paragraph{Optimiser and Hyperparameters.}
All training was performed on a single NVIDIA A10G 24\,GB GPU.
All stages use the Adam optimiser ($\epsilon=10^{-15}$) with position learning rate
$1.6\times10^{-4}$, $\lambda_\text{DSSIM}=0.2$, and densification
percentage $0.01$. No scene-specific hyperparameter tuning was applied; the
same pipeline was run identically across all five scenes.

\section{Results}

\begin{table}[h]
\centering
\small
\setlength{\tabcolsep}{4pt}
\begin{tabular}{lccc}
\toprule
Method / Scene & PSNR\,$\uparrow$ & SSIM\,$\uparrow$ & LPIPS\,$\downarrow$ \\
\midrule
3DGS~\cite{kerbl3Dgaussians} (baseline) & 26.74 & 0.750 & 0.410 \\
Triangle Splat~\cite{Held2025Triangle} (baseline) & 26.43 & 0.757 & 0.359 \\
\midrule
DENSER (Ours)                & & & \\
~~Scene 1                    & 30.016 & 0.7819 & 0.3976 \\
~~Scene 2                    & 29.742 & 0.8028 & 0.3476 \\
~~Scene 3                    & 29.821 & 0.7866 & 0.3359 \\
~~Scene 4                    & 29.514 & 0.8095 & 0.3591 \\
~~Scene 5                    & 30.334 & 0.7747 & 0.3878 \\
\midrule
\textbf{DENSER Mean}         & \textbf{29.885} & \textbf{0.7911} & \textbf{0.3656} \\
\bottomrule
\end{tabular}
\caption{Results on the SoccerNet-NVS challenge test set~\cite{snNVS}. Baseline
numbers are provided by the challenge organisers. Per-scene breakdown shown for DENSER.}
\label{tab:results}
\end{table}

DENSER improves mean PSNR by $\mathbf{+3.15}$\,\textbf{dB} over 3DGS and $\mathbf{+3.46}$\,\textbf{dB} over
Triangle Splat with consistent SSIM gains across all five scenes
(Figure~\ref{fig:comparison}). LPIPS is
marginally higher than Triangle Splat (0.366 vs.\ 0.359), a common trade-off
of ensemble averaging that smooths high-frequency detail while suppressing
structured artefacts.

{
  \small
  \bibliographystyle{ieeenat_fullname_unsorted}
  \bibliography{main}
}

\end{document}